\documentclass[12pt,A4]{article}
\usepackage[margin=1in,footskip=0.25in]{geometry}

\usepackage{lineno,hyperref}

\usepackage{subfig}
\usepackage{graphicx, color}  

\usepackage{algorithm}
\usepackage{algpseudocode}
\usepackage{afterpage}
\usepackage{bm}
\usepackage{amsmath, amsfonts}
\usepackage{multirow}

\usepackage[square,sort,comma,numbers]{natbib}
\usepackage{url}
\usepackage{enumerate}
\usepackage{rotating}

\newcommand{\killpunct}[1]{}

\def\x{\bm{x}}   
 \def\w{\bm{w}}   \def\e{\bm{e}}

 \def\I{\bm{I}}  \def\A{\bm{A}}

 \def\thth{\bm\theta}

\def \LL {\mathcal{L}}

\def \RRR {\mathbb{R}} \def \EEE {\mathbb{E}} 
 
\newcommand{\tabincell}[2]{\begin{tabular}{@{}#1@{}}#2\end{tabular}}  

\begin{document}

	\title{\bf GAMI-Net: An Explainable Neural Network based on Generalized Additive Models with Structured Interactions}
	
	\author{Zebin Yang$^1$, Aijun Zhang$^{1,2}$ and Agus Sudjianto$^2$\\
		{\normalsize  $^1$Department of Statistics and Actuarial Science, The University of Hong Kong,}\\
		{\normalsize Pokfulam Road, Hong Kong}\\
		{\normalsize $^2$Corporate Model Risk, Wells Fargo, USA}}
	\date{}
	
	\maketitle

	\begin{abstract}
			The lack of interpretability is an inevitable problem when using neural network models in real applications. In this paper, an explainable neural network based on generalized additive models with structured interactions (GAMI-Net) is proposed to pursue a good balance between prediction accuracy and model interpretability. GAMI-Net is a disentangled feedforward network with multiple additive subnetworks; each subnetwork consists of multiple hidden layers and is designed for capturing one main effect or one pairwise interaction. Three interpretability aspects are further considered, including a) sparsity, to select the most significant effects for parsimonious representations; b) heredity, a pairwise interaction could only be included when at least one of its parent main effects exists; and c) marginal clarity, to make main effects and pairwise interactions mutually distinguishable. An adaptive training algorithm is developed, where main effects are first trained and then pairwise interactions are fitted to the residuals. Numerical experiments on both synthetic functions and real-world datasets show that the proposed model enjoys superior interpretability and it maintains competitive prediction accuracy in comparison to the explainable boosting machine and other classic machine learning models.

\end{abstract}

\section{Introduction}

Deep learning is one of the leading techniques in artificial intelligence (AI). Despite its great success, a fundamental and unsolved problem is that the working mechanism of deep neural networks is hardly understandable~\citep{montavon2017explaining, lengerich2020purifying}. Without sufficient interpretability, it would be risky to apply these AI systems in real-life applications. A well-trained deep neural network is known to usually have accurate predictive performance on data at hand. However, it may perform abnormally as the data is slightly changed, as its inner decision-making process is unknown. Some recent examples can be referred to as the adversarial attacks, where a convolutional neural network can be easily fooled by its attackers~\cite{tang2019adversarial}.

Interpretable machine learning is an emerging research topic that tries to solve the aforementioned problem and opens up the black-box of complicated machine learning algorithms~\cite{molnar2018interpretable,Murdoch22071}. Two categories of interpretability are generally investigated, i.e., post-hoc interpretability and intrinsic interpretability. In the post-hoc analysis, a fitted model is interpreted using external tools. Examples of this category include the partial dependence plot~(PDP;~\cite{friedman2001greedy}), local interpretable model-agnostic explanations (LIME;~\cite{ribeiro2016should}), SHapley Additive exPlanations (SHAP;~\cite{lundberg2017unified}) and heatmap visualization of deep neural networks~\cite{samek2016evaluating}. In contrast, intrinsic interpretability aims at making the model intrinsically interpretable. A lot of statistical models belong to this category, e.g., the generalized linear model, decision tree, and na\"{i}ve Bayes classifier. In this paper, we limit our focus to the second type of interpretability.

The generalized additive index model (GAIM) is such an intrinsically interpretable model when proper constraints are imposed. It was first proposed by~\cite{friedman1981projection} in the name of projection pursuit regression. GAIM is shown to have close connections with feedforward neural networks~\citep{hwang1994regression}, which has universal approximation capability as the number of hidden nodes is sufficiently large~\citep{hastie2009elements}. The functional relationship between raw features $\bm{x} \in \mathbb{R}^{p}$ and the response $y$ is represented by 
\begin{equation} \label{GAMINet:GAIM}
g(\EEE(y|\x)) = \mu + \sum_{j=1}^{M} h_{j}(\bm{w}_{j}^{T}\bm{x}),
\end{equation}
where $g$ is a pre-specified link function, $\mu$ is the intercept, and $M$ is the number of additive functional components. For each $j=1,\ldots,M$, $\w_j \in\RRR^p$ denotes the projection index and $h_{j}$ is the so-called ridge or nonlinear shape function. Conventionally, GAIM is estimated by the backfitting algorithm, which iteratively estimates a pair of $\{\w_j, h_j\}$ at a time, with other pairs fixed. Nonparametric regression (e.g. smoothing splines) is used to fit the shape functions in (\ref{GAMINet:GAIM}). Such a greedy procedure yields the sub-optimal solution. Recently, GAIM has been reformulated to be an explainable neural network (xNN; \citealp{vaughan2018explainable}). In xNN, a fully-connected multi-layer perceptron is disentangled into a projection layer followed by multiple sub-modular networks, where each subnetwork represents a nonlinear shape function in (\ref{GAMINet:GAIM}). The interpretability of xNN is further enhanced by imposing sparsity, orthogonality, and smoothness constraints~\citep{yang2020enhancing}. As a result of using neural network parametrization of shape functions in xNN, a globally optimal solution can be obtained through full network training. 

The generalized additive model (GAM;~\citealp{hastie1990generalized}) is another intrinsically interpretable model of the form
\begin{equation} \label{GAMINet:GAM}
g(\EEE(y|\x)) = \mu + \sum_{j=1}^{p} h_{j}(x_{j}),
\end{equation}
which is a special case of (\ref{GAMINet:GAIM}) when $M=p$ and $\w_j = \e_j$.  When fitting mathematically complex functions, GAM is less competitive than GAIM, however when solving real problems in specific application domains, GAM is simpler and more interpretable than GAIM.  This difference lies in interpretation of the projection $\bm{z}_{j} = \bm{w}_{j}^{T}\bm{x}$. It is usually difficult to explain the practical meaning of weighted combination of raw features, whereas each raw feature is indeed meaningful. GAM~\citep{hastie1990generalized} deserves to be an important model for interpretable machine learning, due to its intrinsic interpretability of nonlinear main effects. 
An empirical study of GAM based on machine learning datasets was presented by \cite{lou2012intelligible}, which suggested that using tree ensembles to fit nonlinear shape functions in (\ref{GAMINet:GAM}) may achieve better predictive performance than using regression splines. Recently, it draws our attention that \cite{agarwal2020neural} proposed to use neural network representation for the shape functions in GAM, which is the same idea as in xNN~\citep{vaughan2018explainable, yang2020enhancing}.

The interaction effects between individual features can be incorporated into the GAM for performance improvement~\citep{coull2001simple, lou2013accurate, caruana2015intelligible, lengerich2020purifying}. Among them, the GA$^2$M proposed by~\cite{lou2013accurate} is a state-of-the-art extension of (\ref{GAMINet:GAM}) plus pairwise interactions, which is also known as the explainable boosting machine (EBM) with a fast implementation by Microsoft Research team~\citep{nori2019interpretml}. EBM is similar to~\cite{lou2012intelligible} by using tree ensembles to fit either main effect $h_j(x_j)$ or pairwise interaction $f_{jk}(x_j, x_k)$, and it comes with a fast procedure for pairwise interaction detection. It is shown by~\cite{nori2019interpretml} that EBM has the overwhelming prediction performance when compared to some black-box models (including Random Forest, LightGBM, and XGBoost) based on five classification datasets.  

In this paper, a novel xNN structure is proposed by using neural network parametrization for both main effects and pairwise interactions, and we call it GAMI-Net. Unlike EBM based on tree ensembles, we suggest modeling each main effect or pairwise interaction by a fully-connected subnetwork consisting of one or two input nodes, respectively. These subnetworks are then additively combined to form the final output. Each subnetwork can be easily visualized by 1D and 2D plots for the purpose of interpretation. In addition to neural network parametrization, the interpretability of GAMI-Net is enhanced with the following three constraints,
\begin{itemize}
	\item \textbf{Sparsity}. Model parsimony is an essential factor for an interpretable model. In GAMI-Net, only non-trivial main effects and pairwise interactions are included. Pruning of trivial effects is also helpful for reducing the degree of overfitting.
	\item \textbf{Heredity}. The classic heredity principle in statistics is introduced to enhance structural interpretability. That is, a pairwise interaction can only be included in the final model if at least one of its parent main effects is important.
	\item \textbf{Marginal Clarity}. The first two constraints are both employed to select important main effects and pairwise interactions, while marginal clarity serves as a regularization to avoid potential confusion between main effects and their corresponding child pairwise interactions.
\end{itemize}

A three-stage adaptive training algorithm is proposed for GAMI-Net estimation. First, the main effect subnetworks are trained and pruned. Second, important pairwise interactions are selected, fitted, and pruned. Finally, all the important main effects and pairwise interactions are collectively fine-tuned. Numerical experiments on both synthetic functions and real-world datasets are conducted. The results reveal that the fitted GAMI-Net is more interpretable as compared to its benchmark models. Moreover, it is shown that GAMI-Net is competitive regarding predictive performance, which makes it another promising tool for interpretable machine learning.

This paper is organized as follows. Section~\ref{GAMINet:Methodology} presents the proposed GAMI-Net methodology, including the network architecture, the training algorithm, and the interpretability. Two synthetic functions and multiple real-world datasets are used to test the GAMI-Net performance in Section~\ref{GAMINet:Experiments}.

\section{GAMI-Net Methodology} \label{GAMINet:Methodology}
This section first introduces the proposed GAMI-Net architecture, interpretability constraints, and computational algorithm. Further discussions are then provided, regarding the model interpretation and hyperparameter tuning guidelines. Finally, we compare GAMI-Net with its counterpart models from various perspectives.  

\subsection{Network Architecture}
In GAMI-Net, a complex functional relationship is formulated via its lower-order representations, including nonlinear main effects and pairwise interactions\footnote{Higher-order interactions can be treated similarly by GAMI-Net, but for simplicity we focus on pairwise interactions only. Besides, we believe that higher-order interactions are usually rarer than pairwise interactions in practice, and when they exist the interpretation is not as straightforward.}. Let $S_{1}, S_{2}$ denote the set of active main effects and pairwise interactions, respectively. Then, the proposed GAMI-Net is formulated as follows,
\begin{equation} \label{GAMINet:GAMI}
\begin{gathered}
g(\EEE(y|\x)) =  \mu + \sum\limits_{j \in S_{1}} h_{j}(x_{j}) + \sum\limits_{(j, k) \in S_{2}} f_{jk}(x_{j},x_{k}).
\end{gathered}
\end{equation}
Here, each main effect and pairwise interaction is assumed to have zero means, i.e.,
\begin{equation} \label{GAMINet:zero_means}
\begin{gathered}
{\displaystyle  \int h_{j}(x_{j})dF(x_{j})=0, \forall j \in S_{1},} \\
{\displaystyle  \int f_{jk}(x_{j}, x_{k})dF(x_{j}, x_{k})=0, \forall (j, k) \in S_{2},}
\end{gathered}
\end{equation}
where $F(x_{j})$ and $F(x_{j}, x_{k})$ represent the corresponding cumulative distribution functions. To avoid the non-identifiability problem, each pairwise interaction $f_{jk}(x_{j}, x_{k})$ should be nearly orthogonal to its parent main effects $h_{j}(x_{j})$ and $h_{k}(x_{k})$, subject to the marginal clarity constraint (to de discussed below).

\begin{figure*}[!t]
	\centering
	\includegraphics[width=1.0\textwidth]{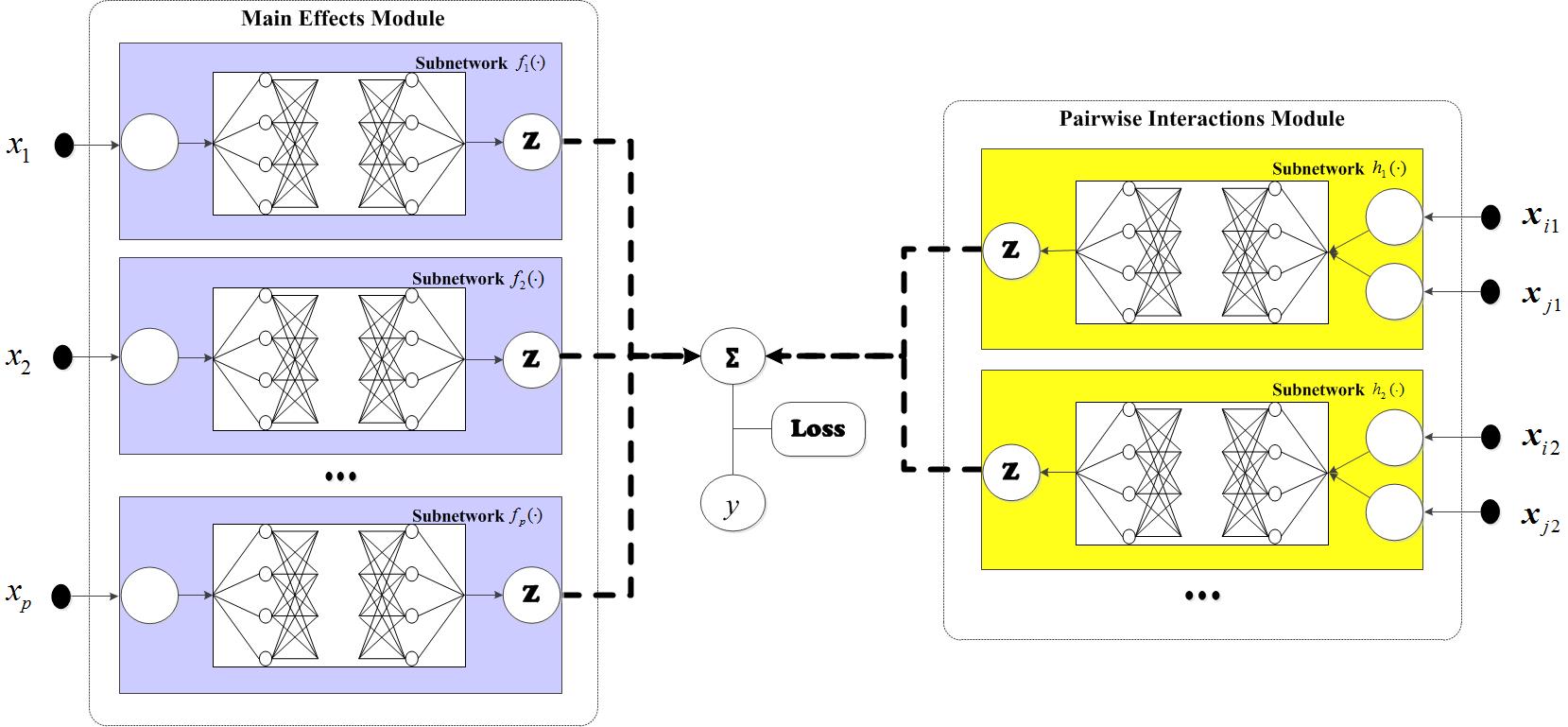}
	\caption{The GAMI-Net architecture. The main effects are first fitted. Then the top-$K$ ranked pairwise interactions are selected and fitted to the residuals, subject to the heredity constraint. The dashed arrows pointing to the $\bm\Sigma$ nodes denote the sparsity constraints, where the trivial subnetworks are pruned. Finally, the marginal clarity is imposed for regularizing pairwise interactions, denoted by the symbol ``C''.}\label{GAMINet:framework}
\end{figure*}

The GAMI-Net architecture is presented in Figure~\ref{GAMINet:framework}. It consists of a main effect module and a pairwise interaction module. Each main effect $h_{j}(x_{j})$ in (\ref{GAMINet:GAMI}) is captured by a subnetwork consisting of one input node, multiple hidden layers, and one output node. Each pairwise interaction $f_{jk}(x_{j}, x_{k})$ in (\ref{GAMINet:GAMI}) is captured a subnetwork with two input nodes. All these networks are linearly combined (plus a bias node for capturing the intercept $\mu$) to produce the final output. More specifically, the main effect subnetwork fits a 1D curve, while the interaction subnetwork approximates a 2D surface. When approximating an arbitrary curve or surface, we can use a single-hidden-layer feedforward neural network with a sufficiently large number of hidden nodes, while modern deep learning training techniques make it feasible to use multiple hidden layers to achieve superior predictive performance. In fact, the multi-layer subnetworks are flexible enough to capture any form of functions upon proper network configuration. Besides, categorical variables are preprocessed using one-hot encoding. The subnetworks used for fitting the main effects of categorical variables can be simplified to many bias nodes, where each node captures the intercept effect of a corresponding dummy variable.

\subsection{Interpretability Constraints}
The proposed GAMI-Net is developed with the sparsity, heredity, and marginal clarity constraints. Specifically, sparsity and heredity constraints are introduced to enhance the interpretability of the fitted model; while the marginal clarity constraint is introduced to make main effects and their child pairwise interactions uniquely identifiable. 

\textbf{Sparsity Constraint}. The principle of parsimony is commonly assumed in statistical machine learning. The sparse models, upon reduction of unnecessary model complexity, not only enjoy computational benefits but also prevent overfitting problems. Moreover, sparsity is an essential building block for model interpretation. For example, a shallow decision tree that uses a few explanatory variables is generally thought to be easily interpretable; however, a deep decision tree involving multiple variables and many leaf nodes can be hardly understandable. For high-dimensional data, the GAM involving all the variables can be too complex to interpret. Therefore, it is critical for GAMI-Net to remove unnecessary main or interaction effects, in order to benefit from efficient computation and enhanced interpretability.

The importance of a main effect or pairwise interaction can be quantified by the variation it explains. Empirically, the variation of the $j$-th main effect can be measured by the sample variance,
\begin{equation} \label{GAMINet:maineffect_contribution}
	D(h_{j})=\frac{1}{n - 1}\sum h_{j}^{2}(x_{j}),
\end{equation}
where $n$ is the sample size. We treat the main effect functions with very small variation as trivial effects, and enforce them to zero, which results in the sparse GAM~\citep{ravikumar2009sparse}. Alternatively, given an integer parameter $s_1$ (between 1 and $p$), GAMI-Net is designed to select the top-$s_1$ main effects ranked by $D(h_{j})$ values, as listed by the index set $S_1$.

Similarly,  the sparsity of pairwise interactions can also be induced by selecting the top-$s_{2}$ pairwise interactions according to the $D(f_{jk})$ value defined by
\begin{equation} \label{GAMINet:interaction_contribution}
D(f_{jk})=\frac{1}{n - 1}\sum f_{jk}^{2}(x_{j}, x_{k}),
\end{equation} 
for all the pairwise interactions. We use the index set $S_{2}$ to denote the list of selected top-$s_{2}$ pairwise interactions. 

\textbf{Heredity Constraint}. In addition to sparsity constraint, hierarchical and hereditary principles are both essential rules for modeling main effects and low-order to high-order interactions. The hierarchical principle states that lower-order effects are generally more important than higher-order effects. The principle of heredity further requires a more strict hierarchical structure between main effects and interactions~\citep{nelder1998selection}, whereas the model violating the heredity principle is thought to be insensible~\citep{mccullagh1983generalized}. The heredity principle has also been used in the variable selection literature~\citep{bien2013lasso, choi2010variable, yuan2009structured, peixoto1987hierarchical}.

There are two versions of the heredity principle, namely strong heredity, and weak heredity. In the case of main effects (indexed by $S_1$) and pairwise interactions (indexed by $S_2$), the strong heredity imposes the constrain that 
$$
\forall  (j,k)\in S_2:  j\in S_1 \ \mbox{ and }\  k\in S_1;
$$
while the weak heredity imposes that
$$
\forall  (j,k)\in S_2:  j\in S_1 \ \mbox{ or }\  k\in S_1. 
$$
That is, a pairwise interaction can be included by $S_2$ only if a) both of its parent main effects are included (strong heredity) by $S_1$, or b) at least one of its parent main effects is included by $S_1$. 


In GAMI-Net, the weak heredity constraint is employed for the following reasons. First, the search space (of pairwise interactions) can be reduced and hence it brings computational efficiency. Second, the resulting model can be improved with enhanced interpretability in the sense of the heredity principle. Third, the heredity principle is empirically supported in statistical modeling literature; see for instance the meta-analysis conducted by~\cite{li2006regularities} for a large number of data sets from published factorial experiments. 

\textbf{Marginal Clarity}. For model identifiability, each main effect or pairwise interaction is assumed to have zero mean in (\ref{GAMINet:zero_means}). However, without further assumptions, the main effects can be easily absorbed by their child interactions and vice versa. There could be multiple representations for a given model, which makes the model estimation unstable and leads to confusion in model interpretation.

The marginal clarity constraint is accordingly introduced to make the model more identifiable.  It is motivated by the functional ANOVA decomposition, in which the original function can be uniquely decomposed into orthogonal components. The weighted functional ANOVA decomposition~\citep{hooker2007generalized} is proposed for handling explanatory variables with empirical distributions, where the orthogonality condition for the $j$-th main effect and corresponding pairwise interaction $(j, k)$ is presented as follows,
\begin{equation}
\int h_{j}\left(x_{j}\right) f_{jk}\left(x_{j}, x_{k}\right) d F(\x) = 0. \label{GAMINet:orthogonality}
\end{equation}
The symbol $F(\x)$ denotes the joint cumulative distribution function. Empirically, the degree of non-orthogonality can be defined by 
\begin{equation}
\Omega(h_{j}, f_{jk}) = \left| \frac{1}{n}\sum h_{j}(x_{j})f_{jk}(x_{j}, x_{k}) \right|. \label{GAMINet:marginal_clarity}
\end{equation}
The smaller the value of $\Omega(h_{j}, f_{jk})$, the more clearly the marginal effect $h_{j}$ is separated from its child interaction $f_{jk}$. The perfect case is when $\Omega(h_{j}, f_{jk})=0$;  in practice, it is acceptable to have $\Omega(h_{j}, f_{jk})$ slightly greater than zero. Hence in GAMI-Net, we penalize the non-orthogonality $\Omega(h_{j}, f_{jk})$ for all $j\in S_1$ and $(j,k)\in S_2$, in a way for pursuing the marginal clarity. Note that a similar interaction purifying method is proposed in~\cite{lengerich2020purifying}, which is based on post-hoc processing and only suitable for piecewise constant functions.

\subsection{Computational Aspects} \label{GAMINet:Estimation}
In this section, we discuss the computational procedures for estimating GAMI-Net. All the unknown parameters in the proposed model are denoted by $\thth$. For each sample $\x$, the prediction is denoted by $\hat{y} = \EEE(y|\x;\thth)$. Combining all the interpretability constraints, GAMI-Net is estimated by solving the following constrained optimization problem, 
	\begin{equation} \label{GAMINet:OptProb}
	\begin{gathered}
	\min_{\thth}\LL_{\lambda}(\thth)  =  l(\thth) + \lambda \sum_{j \in S_{1}} \sum_{(j, k) \in S_{2}} \Omega(h_{j}, f_{jk}),\\
	\mbox{s.t. } {\displaystyle  \int h_{j}(x_{j})dF(x_{j})=0, \forall j \in S_{1},} \\
	{\displaystyle  \int f_{jk}(x_{j}, x_{k})dF(x_{j}, x_{k})=0, \forall (j, k) \in S_{2},}
	\end{gathered}
	\end{equation} 
	where the active sets of main effects and pairwise interactions $S_{1}, S_{2}$ are determined subject to sparsity and heredity constraints. The empirical loss $l(\thth)$ is determined by the type of tasks (e.g. regression or classification). The second term is the marginal clarity regularization, and the regularization strength is denoted by $\lambda \geq 0$.

Referring to Figure~\ref{GAMINet:framework}, an adaptive training algorithm is introduced to sequentially estimate the main effects and pairwise interactions, which can be summarized as the following three stages.
\begin{enumerate}[1)]
	\item Train all the main effect subnetworks for some epochs and prune the trivial main effects according to their contributions and validation performance.
	\item Select at most $K$ pairwise interactions for training and then prune the trivial pairwise interactions according to their contributions and validation performance.
	\item Fine-tune all the network parameters for some epochs.
\end{enumerate}

\textbf{Training Main Effects}. In the first stage, all the main effect subnetworks are simultaneously estimated while the pairwise interaction subnetworks are frozen to zero. The trainable parameters in the network are updated by mini-batch gradient descent with adaptive learning rates determined by the Adam optimizer, which is scalable to very large datasets.

The training will stop as the maximum number of training epochs is reached or the validation performance does not get improved for a certain number of epochs. Each main effect is then centered to have mean zero mean such that the bias node of the output layer represents the overall mean. The trivial main effect subnetworks are then pruned according to the sparsity constraint. Given a null model that only contains the intercept term, we evaluate its performance on the validation set, denoted by $l_{0}$. Then, the most important main effect will be added and we evaluate its validation performance $l_{1}$. Next, the other important main effects will be added one-by-one in the descending order of their contributions (\ref{GAMINet:maineffect_contribution}). The list $\{l_{0}, l_{1}, \cdots, l_{p}\}$ represents their corresponding validation losses.

In general, when more and more main effects are added, the validation loss would show a decreasing trend. However, including too many main effects can lead to overfitting, as reflected by the turning trend in the loss curve. According to the sparse modeling principle, those main effects after the turning point should be pruned. In practice, a tolerance threshold $\eta$ is introduced to balance the level of sparsity and predictive performance. That is, $s_{1}$ is set to the minimal index whose validation loss is smaller than or equal to $(1 + \eta) \min \{l_{0}, l_{1}, \cdots, l_{p}\}$. The active set $S_{1}$ is determined as the list of top-$s_{1}$ important main effects.

\textbf{Training Interactions}. After the top-$s_1$ important main effects are captured, the next step is to train the pairwise interaction subnetworks. In total there exist $p(p-1)/2$ possible pairwise interactions that can be tested, which is extremely time-consuming especially for a large $p$. According to the weak heredity constraint, we consider those candidates of pairwise interactions with at least one of its parent main effects belonging to $S_1$. This reduces the computational complexity a lot when $s_1$ is much less than $p$. Besides, an interaction filtering procedure is introduced to remove the pairwise interactions which are less likely to be important. There exist many interaction detection methods in the literature, to list a few, additive Grove~\citep{sorokina2007additive}, RuleFit~\citep{friedman2008predictive}, hierarchical Lasso~\citep{bien2013lasso}, neural network-based interaction detection~\cite{tsang2018detecting}, and shallow tree-like model-based pairwise interaction ranking~\citep{lou2013accurate}. 

In GAMI-Net, we employ the interaction ranking algorithm proposed in~\cite{lou2013accurate}, subject to the heredity constraint. The modified pairwise interaction filtering algorithm selects the top-$K$ pairwise interactions through the following steps.
\begin{enumerate}[1)]
	\item Obtain the prediction residuals from the main effect training stage.
	\item For each $j < k$ with $j \in S_{1}$ or $k \in S_{1}$, evaluate the strength of interaction $(j, k)$ by building shallow tree-like models between variables $(x_{j}, x_{k})$ and the residuals; the strength of interaction $(j, k)$ is set to the minimal fitting loss across all evaluated trees~\citep{lou2013accurate}.
	\item Rank all the evaluated pairwise interactions and obtain the top-$K$ pairwise interactions.
\end{enumerate}
Next, the selected top-$K$ pairwise interactions are simultaneously trained using the mini-batch gradient algorithm, subject to the marginal clarity regularization. Note that in this stage, the main effect subnetworks are fixed. Each estimated pairwise interaction is centered to have mean zero, for which the offset is added to the bias node in the output layer. 

Pruning of pairwise interaction effects is similar to that of the main effects. We start from the pre-trained model with the intercept term and active main effects. The top-ranked pairwise interactions are sequentially added to the model, together with the record of their corresponding validation losses. For simplicity, we use the same tolerance threshold $\eta$ as for main effect pruning, in order to balance the level sparsity and predictive performance.  Thus, $s_{2}$ can be determined accordingly, and the active set   $S_{2}$ is formed as the list of the top-$s_{2}$ important pairwise interactions.

\textbf{Fine Tuning}. The first two stages perform a structured variable selection. In the final stage, a fine-tuning procedure is implemented to jointly retrain all the active subnetworks, where the marginal clarity regularization is still imposed between main effects and pairwise interactions. All the main effects and pairwise interactions are re-centered. We find such a fine-tuning step is helpful to solve the following two problems: 
\begin{enumerate}[1)]
	\item The removal of trivial main effects or pairwise interactions may lead to biased estimation;
	\item The pairwise interactions estimated separately are conditional on the pre-trained main effects (subject to marginal clarity regularization), which can limit the predictive performance. 
\end{enumerate}
These two problems can be mitigated via jointly retraining all the selected main effects and pairwise interactions so that the predictive performance of GAMI-Net can be further improved.

\subsection{Interpretability of GAMI-Net}
The proposed GAMI-Net is intrinsically interpretable in the following aspects.

\textbf{Importance Ratio (IR)}. The an estimated model by GAMI-Net, we can inspect the contribution of each individual variable to the overall prediction. The IR of each main effect can be quantitatively measured by 
\begin{equation}\label{GAMINet:IR_main}
\begin{gathered}
\mbox{IR}(j) = D(h_{j}) / T,
\end{gathered}
\end{equation} 
where $T = \sum_{j \in S_{1}}D(h_{j}) + \sum_{(j, k) \in S_{2}}D(f_{jk})$. Similarly, the IR of each pairwise interaction can be measured by
\begin{equation}\label{GAMINet:IR_interaction}
\begin{gathered}
\mbox{IR}(j, k) = D(f_{jk}) / T.
\end{gathered}
\end{equation} 
The $\mbox{IR}$'s of all the effects sum up to one. In practice, we can sort the effect importance according to the IR values in descending order. The effects of large IR values are more important.

The definition of IR is related to that of Sobol indices~\citep{sobol2001global}. The main difference lies in that Sobol indice is derived under the assumption that all the variables are independent and uniformly distributed; while IR is based on the empirical distributions of explanatory variables.

\textbf{Global Interpretation}. In addition to measuring the importance of each estimated effect, we can further inspect the relationship between one/two individual variables and the response by visualizing the fitted shape functions. Unlike the post-hoc diagnostic tool PDP~\citep{friedman2001greedy}, the partial dependence relationships can be directly obtained from GAMI-Net. We suggest using the 1D line plots for numerical variables and the bar charts for categorical variables to show the input-output relationship, which can be linear, convex, monotonic, and any other forms. These plots can be directly drawn based on the final estimates of $h_j(x_j)$ for $j\in S_1$. Moreover, we suggest using the 2D heatmap for visualizing each estimated pairwise interaction, which shows the joint effect of the two underlying variables. See e.g. Figure~\ref{GAMINet:S1_Visu} (middle panel) for such kinds of plots.

\textbf{Local Interpretation}. The prediction by GAMI-Net is also easy to be locally explained, leading to a transparent decision-making system. Given a sample $\x$, the model not only outputs the final decision but also the concrete function form (\ref{GAMINet:GAIM}) with the input $\x$. The values of each additive component, i.e. marginal main or pairwise interaction effects, can be directly obtained. These marginal effects can be rank-ordered for understanding the decision for the input $\x$ specifically. Besides, the sensitivity of prediction to small changes of an explanatory variable can be quantitatively investigated by the corresponding 1D line plots (or bar charts) and 2D heatmaps.

\subsection{Hyperparameters} \label{GAMINet:hyperparameters}
Some hyperparameters for GAMI-Net can be configured with the following default settings (for numerical experiments in the next section). The maximal number of pairwise interactions is set to $K = 20$. For simplicity, each subnetwork is configured to have 5 ReLU hidden layers each with 40 nodes.  It is worth mentioning that the choice of activation will affect the resulting functional forms of the fitted model. Using ReLU, the fitted curves are piecewise linear; while using hyperbolic tangent, the fitted curves can be more smooth. 

The subnetwork weights are initialized using the Gaussian orthogonal initializer. The initial learning rate of the Adam optimizer is set to 0.0001. The numbers of training epochs for the three training stages are set to 5000, 5000, and 500, respectively. The mini-batch sample size is determined according to the sample sizes of different datasets. A 20\% validation set is split for early stopping, and the early stopping threshold is set to be 50 epochs. The tolerance threshold $\eta$ is set to be 1\% of the minimal validation loss. The marginal clarity regularization strength $\lambda$ can be empirically selected from 0.0001 to 1.

Finally, a demo implementation of the proposed GAM-Net is publicly available, which can be found on the Github\footnote{https://github.com/ZebinYang/gaminet}, which also includes the numerical examples presented in this paper. This package is based on the \textsl{TensorFlow 2.0} platform using the Python language. 

\subsection{Comparison with Related Methods}
Unlike traditional spline-based GAMs, GAMI-Net uses neural networks to model the non-parametric shape functions. The proposed GAMI-Net is also closely related to the explainable boosting machine (EBM;~\citealp{lou2013accurate}), as both of them are based on main effects and pairwise interactions. 

In EBM, each main effect or pairwise interaction is estimated via gradient boosted shallow trees, which is modified from the standard gradient boosting model~\citep{friedman2001greedy}. Therefore, the estimated shape functions by EBM are all piecewise constant. Empirically, gradient boosted shallow trees are shown to have strong approximation ability, which makes EBM even comparable to black-box models like random forest and neural networks~\citep{lou2012intelligible, chang2020interpretable}. Despite its predictive performance, EBM sometimes outputs shape functions with unexpected jumps, which are hard to explain. Such a problem may become worse when there exist outliers or noisy samples.

In contrast, spline-based GAMs and neural network-based GAMI-Net usually output continuous shape functions for numerical variables. Using splines, the smoothness of the fitted functions can be partially controlled by the choice of spline orders; while for GAMI-Net, the fitted functions can be piecewise linear (e.g., when using ReLU) or more smooth (e.g., when using sigmoid). Such continuous or smooth shape functions can prevent unexpected jumps and therefore warrant the model interpretability.

%

The proposed GAMI-Net tends to be more efficient and more interpretable than EBM. In EBM, all the main effects are included in the final model; the number of active pairwise interactions in EBM can only be pre-specified (this is not flexible) or tuned by cross-validation (this is time-consuming). The resulting model from EBM can be extremely complex for high-dimensional data. Besides, without marginal clarity constraint in EBM, one main effect and its corresponding child pairwise interactions may be mutually absorbed, which leads to non-identifiable results. Such problems can be well addressed by the interpretability constraints imposed on GAMI-Net, including sparsity, heredity, and marginal clarity.

\section{Numerical Experiments}\label{GAMINet:Experiments}
In this section, the proposed GAMI-Net is tested on a synthetic example and an extensive list of real-world datasets. 

\subsection{Experimental Setup}	
Several benchmark models are included for comparison, including EBM, spline-based GAM, generalized linear models~(GLM),  multi-layer perceptron~(MLP), random forest~(RF), and extreme gradient boosting~(XGBoost). Specifically, EBM is implemented by the open-source Python package~\textsl{interpret}~\citep{nori2019interpretml}. The spline-based GAM is based on the implementation of the pyGAM package~\citep{serven2018pygam}, and we use pyGAM to denote spline-based GAM in the remaining part of this paper. For the other benchmarks, GLM, MLP, and RF are all available in the \textsl{Scikit-learn} package, and XGBoost is implemented by the \textsl{xgboost} package. In particular, GLM uses Lasso for regression tasks and $\ell_1$-shrinkage logistic regression for binary classification tasks. The comparative results are grouped into two categories, intrinsically interpretable models (GAMI-Net, EBM, pyGAM, and GLM) and black-box models (MLP, RF, and XGBoost).

By default, we split a dataset into training (80\%) and test (20\%) sets upon random permutation. For hyperparameter tuning, a 20\% hold-out validation set is further split from the training set. GAMI-Net is configured and trained using the settings described in Section~\ref{GAMINet:hyperparameters}. By default, the strength of the marginal clarity regularization is set to 1 in the simulation study and 0.1 in all the real-world datasets. The rationale of such settings is further justified through ablation studies. In EBM, the number of interactions is set to 20, and all the other hyperparameters are set to the default values. In pyGAM, the smoothness regularization strength is tuned within the package's recommended range. The $\ell_{1}$ regularization strength for Lasso (or logistic regression) is tuned within $\{10^{-2},10^{-1},10^{0},10^{1},10^{2}\}$. For black-box MLP, the hidden layer architecture is set to $[40, 20]$ with hyperbolic tangent nodes. Finally, the number of base estimators is set to 500 for both RF and XGBoost; and for each of them, the maximum tree depth is tuned within $ \{3, 4, 5, 6, 7, 8\}$. The predictive performance is measured by the root mean squares error (RMSE) for regression tasks and the area under the ROC curve (AUC) for binary classification tasks. All the experiments are repeated 10 times and we report the average results.

\subsection{Simulation Study}
A synthetic function is used to demonstrate the proposed method, in which both main effects and pairwise interactions are included, as follows,
	\begin{equation}
	\begin{aligned}
	y =  & 8\left(x_{1}-\frac{1}{2}\right)^{2} + \frac{1}{10}e^{\left(-8x_{2}+4\right)} + 3\sin{(2 \pi x_{3}x_{4})}+ \\ 
	& 5e^{-2(2x_{5}-1)^{2} - \frac{1}{2}\left[15x_{6} + 12(2x_{5}-1)^{2} - 13\right]^{2}} + \varepsilon,
	\end{aligned} \label{GAMINet:simu}
	\end{equation}
	where the response is calculated via complicated nonlinear transformations of the explanatory variables plus a noise term generated from the standard normal distribution. In addition to ($x_{1}, \cdots, x_{6}$), a large number of noisy variables ($x_{7}, \cdots, x_{100}$) are also introduced, which has no contribution to the response. These explanatory variables are generated within the domain $[0, 1]^{100}$, with 3 different distributions, i.e, uniform distribution $\rm U(0, 1)$, normal distribution $\rm N(0.5, 0.2^2)$ truncated within $[0, 1]$, and exponential distribution $\rm Exp(0.5)$ truncated within $[0, 1]$. For each of these three distributions, four different sample sizes are tested, i.e., $n=\{1000, 2000, 5000, 10000\}$.

	\begin{sidewaystable}[!t]
	\centering
	\small
	\renewcommand\tabcolsep{3pt}
	\renewcommand\arraystretch{1.2}
	\caption{Testing RMSE comparison of the synthetic function.} \label{GAMINet:Simu}
	\begin{tabular}{cc|cccc|ccc}
		\hline
		Distribution &  $n$  &          GAMI-Net          &       EBM       &      pyGAM      &       GLM       &       MLP       &       RF        &          XGBoost           \\ \hline
		uniform    & 1000  & $\mathbf{1.712}$$\pm$0.403 & 2.587$\pm$0.000 & 2.998$\pm$0.138 & 2.999$\pm$0.000 & 3.201$\pm$0.000 & 2.458$\pm$0.160 & $\mathbf{2.383}$$\pm$0.172 \\
		uniform    & 2000  & $\mathbf{1.242}$$\pm$0.205 & 2.529$\pm$0.000 & 2.889$\pm$0.069 & 3.124$\pm$0.000 & 3.054$\pm$0.000 & 2.127$\pm$0.056 & $\mathbf{2.077}$$\pm$0.099 \\
		uniform    & 5000  & $\mathbf{1.081}$$\pm$0.053 & 2.158$\pm$0.000 & 2.456$\pm$0.058 & 3.003$\pm$0.000 & 2.705$\pm$0.000 & 1.895$\pm$0.058 & $\mathbf{1.790}$$\pm$0.045 \\
		uniform    & 10000 & $\mathbf{1.044}$$\pm$0.019 & 1.799$\pm$0.132 & 2.450$\pm$0.046 & 3.103$\pm$0.065 & 2.615$\pm$0.057 & 1.822$\pm$0.043 & $\mathbf{1.634}$$\pm$0.013 \\
		normal    & 1000  & $\mathbf{1.842}$$\pm$0.352 & 2.029$\pm$0.000 & 2.528$\pm$0.120 & 2.399$\pm$0.000 & 2.484$\pm$0.000 & 1.914$\pm$0.104 & $\mathbf{1.882}$$\pm$0.121 \\
		normal    & 2000  & $\mathbf{1.385}$$\pm$0.177 & 1.914$\pm$0.000 & 2.526$\pm$0.081 & 2.579$\pm$0.000 & 2.474$\pm$0.000 & 1.702$\pm$0.052 & $\mathbf{1.660}$$\pm$0.066 \\
		normal    & 5000  & $\mathbf{1.092}$$\pm$0.118 & 1.854$\pm$0.000 & 2.105$\pm$0.050 & 2.505$\pm$0.000 & 2.148$\pm$0.000 & 1.541$\pm$0.059 & $\mathbf{1.487}$$\pm$0.042 \\
		normal    & 10000 & $\mathbf{1.043}$$\pm$0.020 & 1.647$\pm$0.124 & 2.010$\pm$0.032 & 2.560$\pm$0.055 & 2.110$\pm$0.024 & 1.485$\pm$0.027 & $\mathbf{1.358}$$\pm$0.026 \\
		exponential  & 1000  & $\mathbf{1.408}$$\pm$0.049 & 2.066$\pm$0.000 & 2.357$\pm$0.137 & 2.336$\pm$0.000 & 2.519$\pm$0.000 & 2.010$\pm$0.107 & $\mathbf{1.960}$$\pm$0.098 \\
		exponential  & 2000  & $\mathbf{1.264}$$\pm$0.108 & 1.965$\pm$0.000 & 2.221$\pm$0.083 & 2.360$\pm$0.000 & 2.325$\pm$0.000 & 1.890$\pm$0.101 & $\mathbf{1.815}$$\pm$0.069 \\
		exponential  & 5000  & $\mathbf{1.062}$$\pm$0.044 & 1.833$\pm$0.000 & 1.954$\pm$0.045 & 2.508$\pm$0.000 & 2.303$\pm$0.000 & 1.767$\pm$0.056 & $\mathbf{1.624}$$\pm$0.040 \\
		exponential  & 10000 & $\mathbf{1.027}$$\pm$0.019 & 1.614$\pm$0.104 & 1.893$\pm$0.023 & 2.523$\pm$0.043 & 2.078$\pm$0.030 & 1.659$\pm$0.037 & $\mathbf{1.520}$$\pm$0.034 \\ \hline\hline
	\end{tabular}
\end{sidewaystable}

Table~\ref{GAMINet:Simu} reports the averaged test set RMSE and standard deviation (over 10 repetitions) of different models on this synthetic dataset, respectively. For each setting, the best interpretable and black-box models are both highlighted in bold, respectively. It can be observed that the proposed GAMI-Net outperforms all the compared models including both interpretable and black-box models. In all the tested cases, GAMI-Net outperforms the black-box models including MLP and RF.

\begin{figure}[!t]
	\centering
	\includegraphics[width=1.0\textwidth]{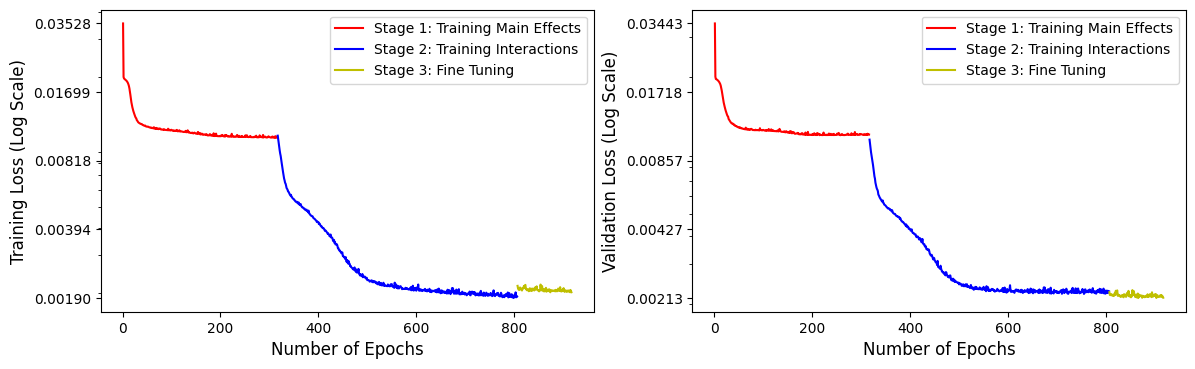}
	\caption{The training and validation trajectories of GAMI-Net for the synthetic function (uniform distribution; $n=10000$).}\label{GAMINet:s1_traj}	
	
	\vspace{0.5cm}
	\centering
	\includegraphics[width=1.0\textwidth]{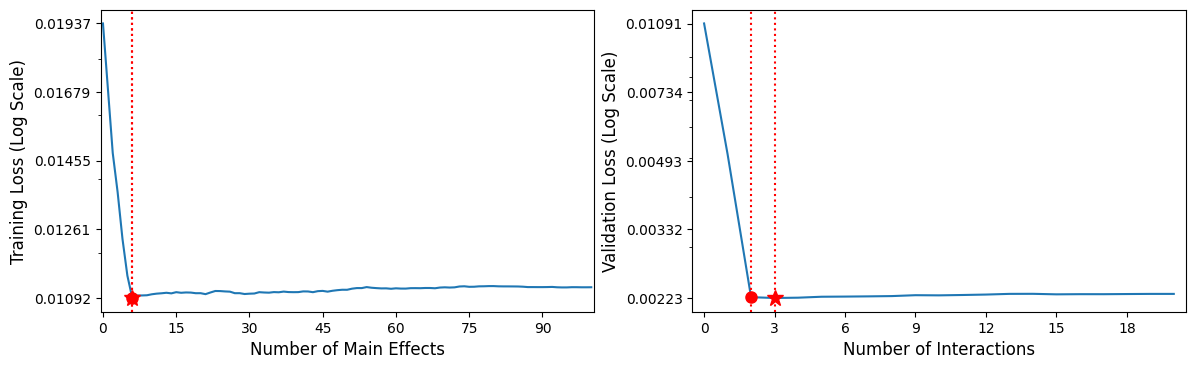}
	\caption{The validation loss for determining $s_{1}, s_{2}$ for the synthetic function (uniform distribution; $n=10000$).}\label{GAMINet:s1_regu}	
\end{figure}

The training and validation losses of GAMI-Net is presented in Figure~\ref{GAMINet:s1_traj}, for the case with uniform distribution and $n=10000$. It can be observed that the losses decrease significantly as pairwise interactions are added to the network, which shows the necessity of adding pairwise interactions to GAM. At the beginning of the fine-tuning stage, there exists a sudden jump of training losses (increase) and validation losses (decrease), which corresponds to the pruning of trivial pairwise interactions. Besides, the validation losses for determining the optimal number of main effects and pairwise interactions are visualized in Figure~\ref{GAMINet:s1_regu}. The left and right x-axises denote the number of included main effects and pairwise interactions, respectively. The optimal number of main effects / pairwise interactions is marked by red star symbols, and the red circle symbols denote the solutions that are acceptable due to the sparsity consideration. The results show that $s_{1} = 6$ main effects and $s_{2} = 2$ pairwise interactions are included in GAMI-Net. The marginal benefits of adding more effects could be extremely small and may even lead to the overfitting problem. 

\begin{figure*}[!t]
	\centering
	\subfloat[Ground Truth]{
		\label{GAMINet:Simu1_GT} 
		\includegraphics[width=0.9\textwidth]{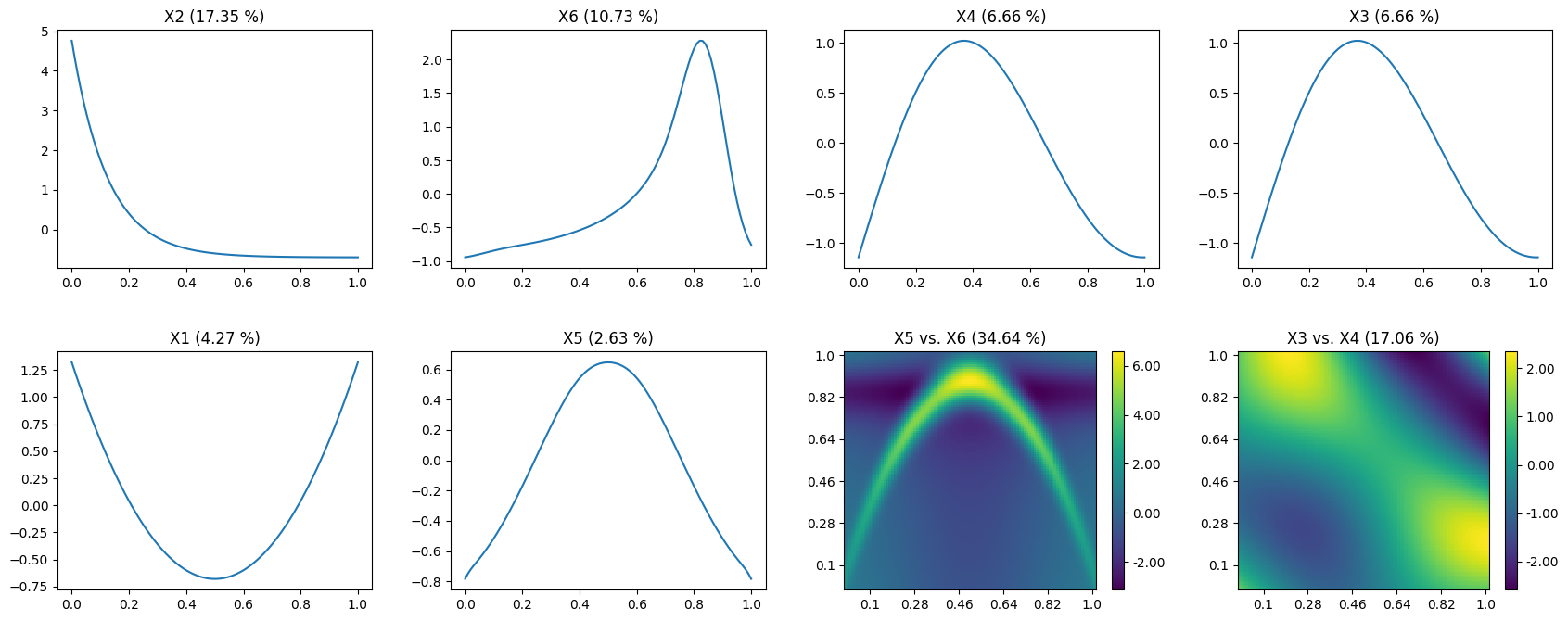}}\\
	\subfloat[GAMI-Net]{
		\label{GAMINet:Simu1_GAMI} 
		\includegraphics[width=0.9\textwidth]{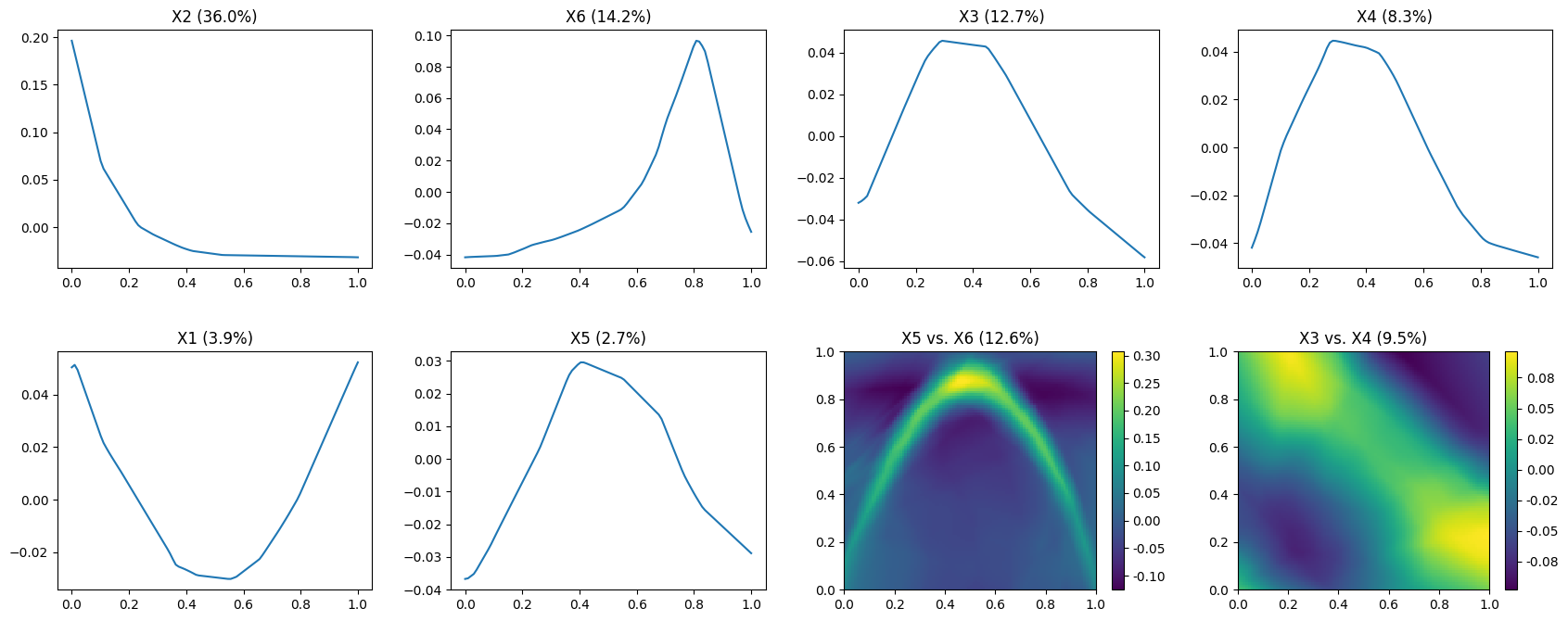}} \\
	\subfloat[EBM]{
		\label{GAMINet:Simu1_EBM} 
		\includegraphics[width=0.9\textwidth]{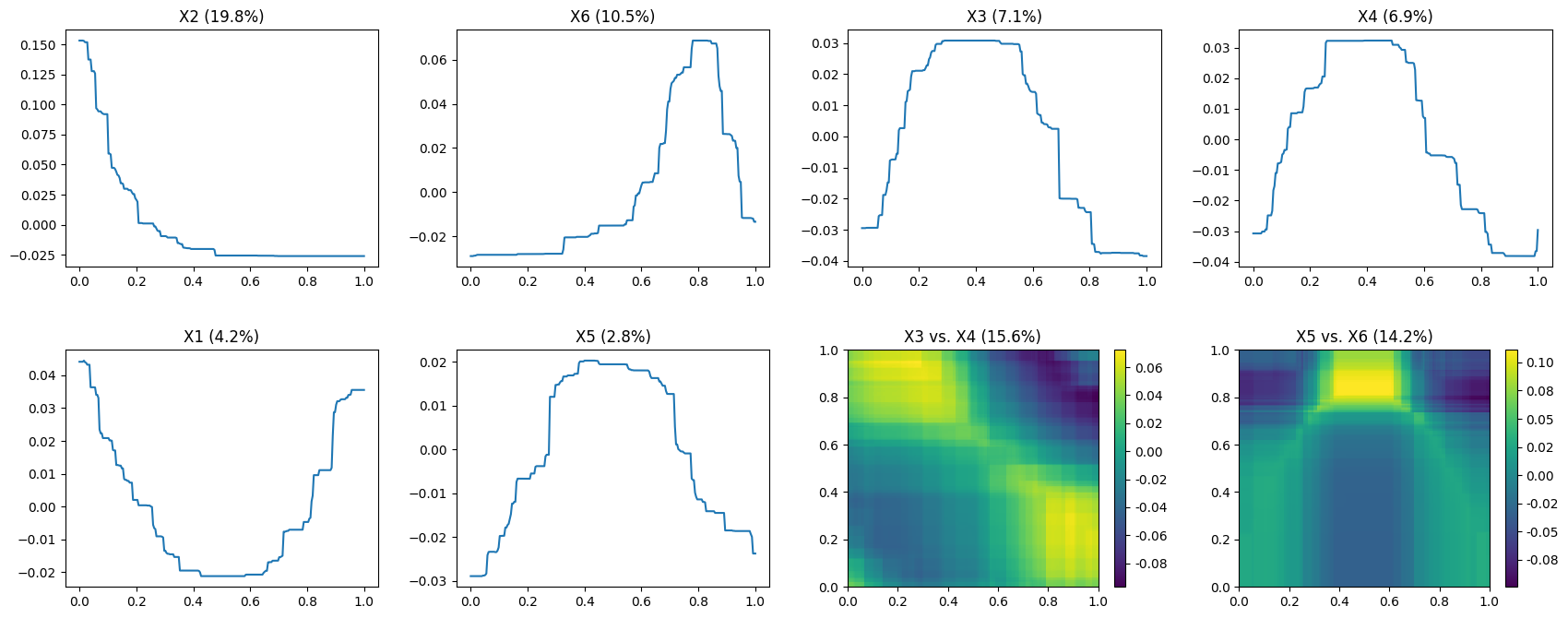}} 
	\caption{The fitted results of GAMI-Net and EBM vs. the ground truth of the synthetic function (uniform distribution; $n=10000$).}\label{GAMINet:S1_Visu}	
\end{figure*}

Figures~\ref{GAMINet:Simu1_GT} and \ref{GAMINet:Simu1_GAMI} draw the ground truth and global interpretation of GAMI-Net (with uniform distribution and $n=10000$). Note the original formulation (\ref{GAMINet:simu}) has only 2 active main effects $(x_{1}, x_{2})$ and 2 active interaction effects $\{(x_{3},x_{4}), (x_{5},x_{6})\}$. But according to the functional ANOVA decomposition, this formula can be rewritten such that the marginal main effects are extracted from the interactions. Therefore, the active main effects also include $x_{3}, x_{4}, x_{5}, x_{6}$. Each main effect / pairwise interaction is ranked in the descending order of IR, and the pairwise interactions are all presented behind main effects. It can be observed that all the 6 main effects and 2 pairwise interactions are successfully captured by GAMI-Net, which is close to that of the ground truth. 

Since EBM does not have a pruning procedure, the final model of EBM includes 100 main effects and 20 pairwise interactions. To make a valid comparison, we also draw its first 6 main effects and first 2 pairwise interactions in Figure~\ref{GAMINet:Simu1_EBM}. The results indicate that EBM can also approximately capture the shape of these important effects. However, due to the use of gradient boosting trees, the estimated shape functions are all piecewise constant and the existence of sudden jumps makes it hard to interpret. Second, we also calculate IR for each effect in EBM using the same method as in GAMI-Net. The result of EBM is shown to have a larger bias as compared to the actual model. For example, the interaction $(x_{5}, x_{6})$ is underestimated and the overall IR captured by these true effects is just around 80\%. That means the noise effects take more than 10\% of the contribution. 

\begin{table}
	\centering
	\renewcommand\tabcolsep{3pt}
	\renewcommand\arraystretch{1.2}
	\caption{Comparison results of the synthetic function (uniform distribution; $n=10000$).} \label{GAMINet:Simu_Ablation}
	\begin{tabular}{c|c|cccc}
		\hline
		\multicolumn{2}{c|}{Model}                                 &   Train RMSE    &    Val RMSE     &    Test RMSE    &   Clarity Loss    \\ \hline
		\multicolumn{2}{c|}{XGBoost}                                & 0.135$\pm$0.103 & 1.630$\pm$0.046 & 1.634$\pm$0.013 &         -         \\
		\multicolumn{2}{c|}{RF}                                  & 1.500$\pm$0.019 & 1.840$\pm$0.060 & 1.822$\pm$0.043 &         -         \\
		\multicolumn{2}{c|}{MLP}                                  & 2.320$\pm$0.076 & 2.372$\pm$0.119 & 2.615$\pm$0.057 &         -         \\
		\multicolumn{2}{c|}{GLM}                                  & 3.062$\pm$0.039 & 3.123$\pm$0.094 & 3.103$\pm$0.065 &         -         \\
		\multicolumn{2}{c|}{pyGAM}                                 & 2.253$\pm$0.025 & 2.459$\pm$0.057 & 2.450$\pm$0.046 &         -         \\
		\multicolumn{2}{c|}{EBM}                                  & 1.155$\pm$0.157 & 1.211$\pm$0.165 & 1.799$\pm$0.132 & 0.0007$\pm$0.0001 \\ \hline
		\multirow{6}{*}{\shortstack{G\\A\\M\\I\\N\\E\\T}} &               $\lambda=10^{0}$                & 0.999$\pm$0.012 & 1.053$\pm$0.023 & 1.044$\pm$0.019 & 0.0003$\pm$0.0001 \\
		&               $\lambda=10^{-1}$               & 0.970$\pm$0.011 & 1.061$\pm$0.029 & 1.058$\pm$0.027 & 0.0002$\pm$0.0001 \\
		&               $\lambda=10^{-2}$               & 0.966$\pm$0.013 & 1.059$\pm$0.021 & 1.056$\pm$0.024 & 0.0006$\pm$0.0004 \\
		&               $\lambda=10^{-3}$               & 0.959$\pm$0.016 & 1.069$\pm$0.020 & 1.058$\pm$0.018 & 0.0024$\pm$0.0013 \\
		&               $\lambda=10^{-4}$               & 0.967$\pm$0.019 & 1.079$\pm$0.025 & 1.072$\pm$0.031 & 0.0037$\pm$0.0025 \\ \cline{2-6}
		& \tabincell{c}{No Heredity\\ $\lambda=10^{0}$} & 1.008$\pm$0.010 & 1.054$\pm$0.019 & 1.045$\pm$0.022 & 0.0002$\pm$0.0001 \\ \hline\hline
	\end{tabular}
\end{table}

The benefits of introducing the sparsity constraint in GAMI-Net is already demonstrated in Figure~\ref{GAMINet:s1_regu}. Moreover, ablation studies are conducted to justify the use of heredity and marginal clarity constraints. In Table~\ref{GAMINet:Simu_Ablation}, we report the training, validation, and test RMSE of the benchmark models. There exist several reasons that EBM fails in this task. First, as the ground truth function is continuous, the piecewise constant fits cannot well capture the ground truth. Second, due to the lack of sparsity consideration, EBM suffers from the overfitting problem. Third, as the main effects are not well captured, the correct pairwise interactions may not always be detected.

Then, the results of GAMI-Net with different marginal clarity regularization strengths are also reported in Table~\ref{GAMINet:Simu_Ablation}. The last row denotes the results of GAMI-Net without heredity constraint and $\lambda=10^{0}$. The marginal clarity losses are calculated for both EBM and GAMI-Net, via 
	\begin{equation}
	\begin{aligned}
	\sum_{j \in S_{1}} \sum_{(j, k) \in S_{2}} \Omega(h_{j}, f_{jk}),
	\end{aligned}
	\end{equation} 
	which is the smaller, the better. The results show that the increase of $\lambda$ can help a) prevent from overfitting, according to the validation RMSE; b) make the model more identifiable, see the decreasing trend of marginal clarity loss. As is discussed in the previous sections, the purpose of using heredity constraint is to help reduce the search space of interactions and make the model structurally more interpretable. The possible drawback of heredity lies in it may reduce the predictive performance. However, it is observed that the inclusion of heredity constraint does not have a significant influence on the predictive performance. Therefore, the use of heredity constraint can be viewed as a bonus term for GAMI-Net.

\subsection{Bank Marketing Dataset}
This dataset is typically used in a binary classification setting (\url{https://archive.ics.uci.edu/ml/datasets/Bank+Marketing}). 
It has 45211 samples with 9 categorical variables and 7 numerical variables, denoting a client's age, education, job, and other related information. The goal is to predict whether a client will subscribe to the term deposit.

\begin{figure*}[!t]
	\centering
	\subfloat[GAMI-Net]{
		\label{GAMINet:bank_gaminet} 
		\includegraphics[width=0.475\textwidth]{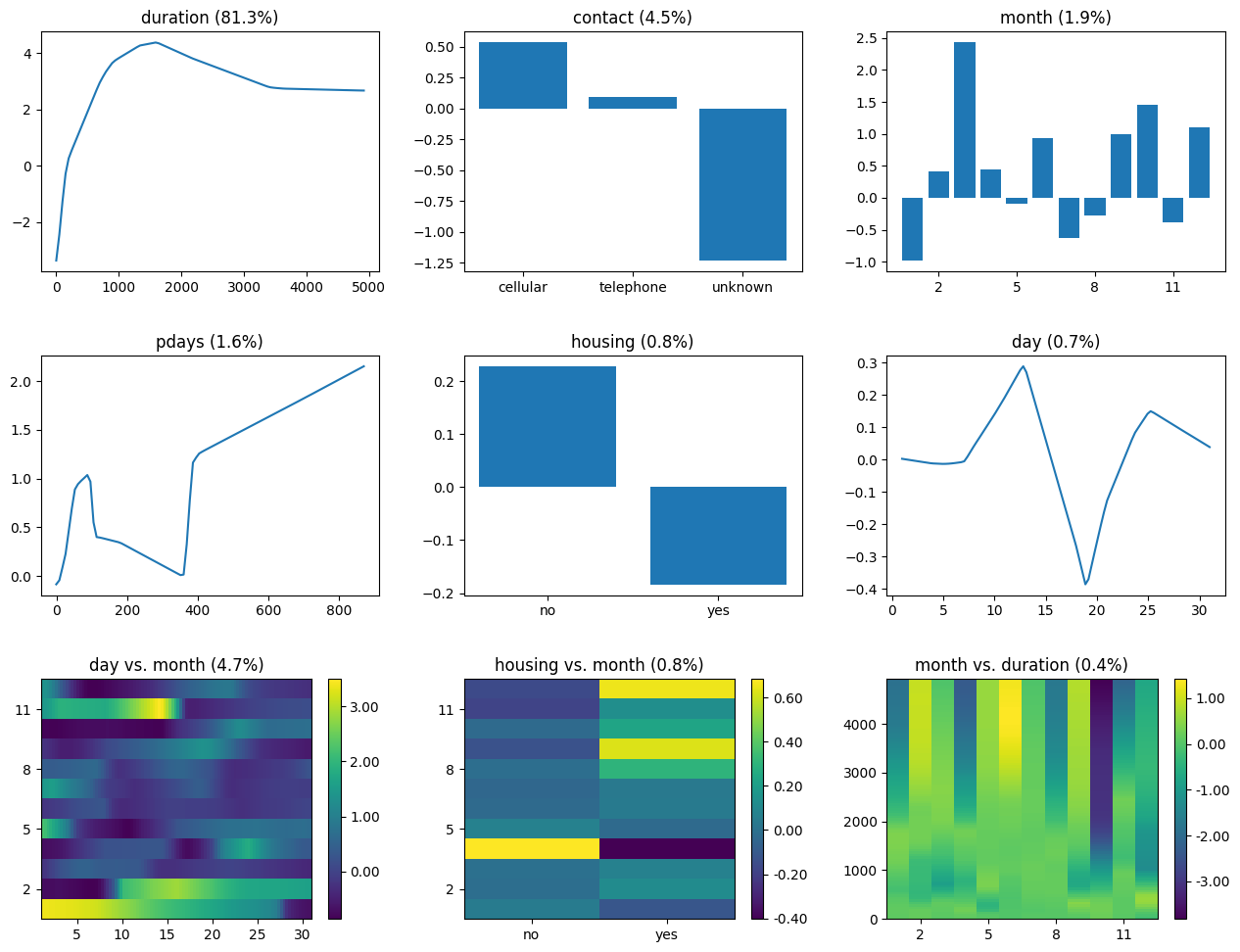}}
	\quad
	\subfloat[EBM]{
		\label{GAMINet:bank_ebm} 
		\includegraphics[width=0.475\textwidth]{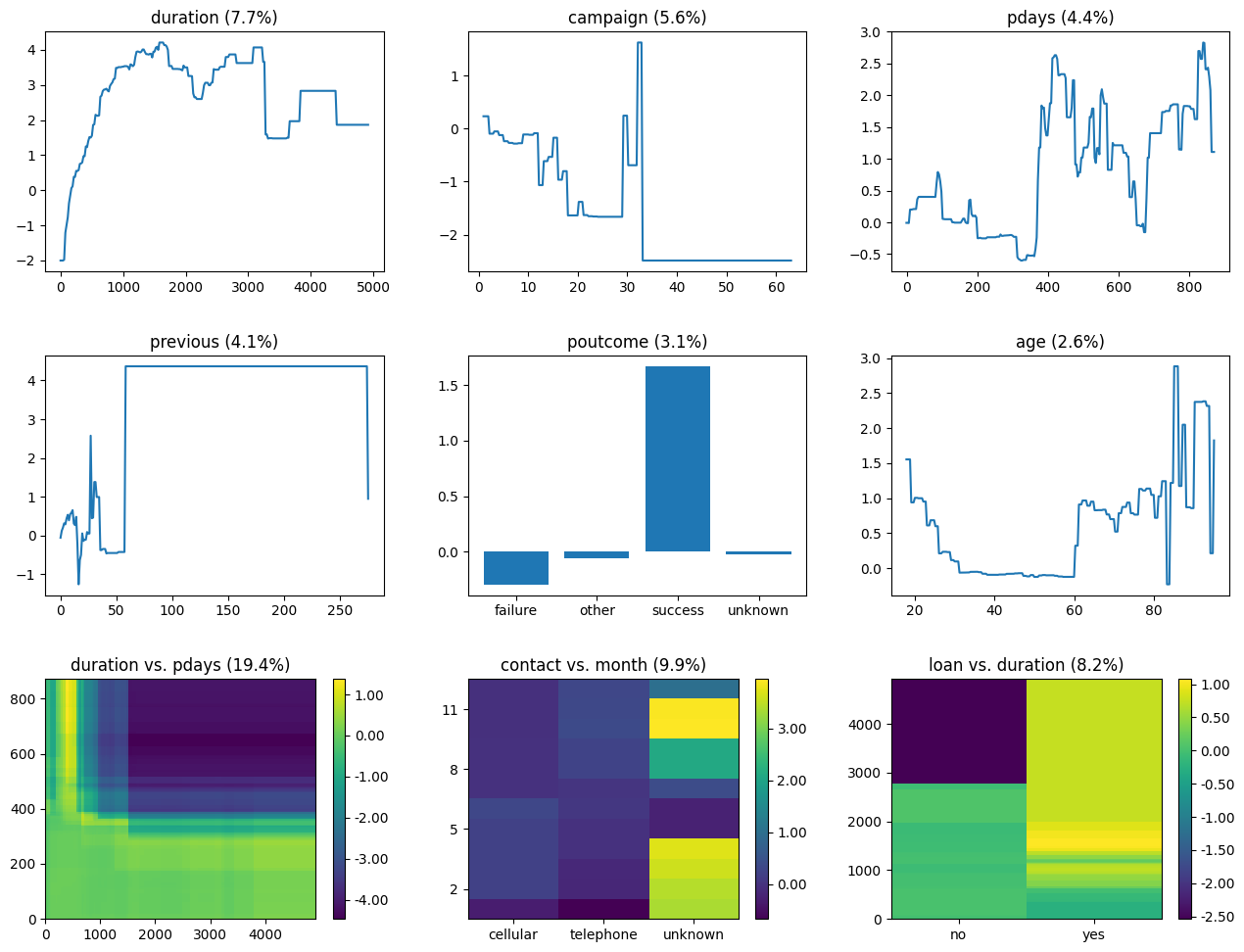}} 
	\caption{GAMI-Net vs. EBM global interpretation for the bank marketing dataset.}\label{GAMINet:bank_global}	
\end{figure*}

\begin{figure}[!t]
	\centering
	\includegraphics[width=0.8\textwidth, height=0.35\textheight]{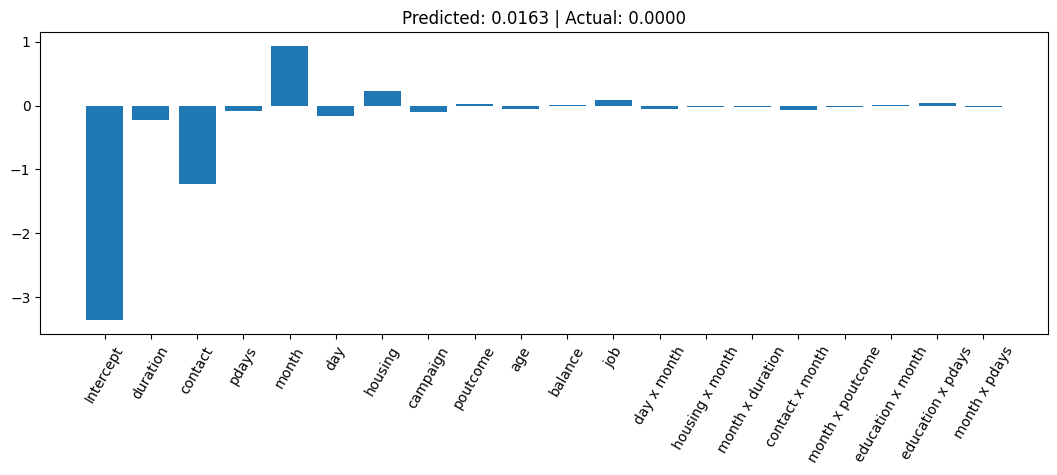}
	\caption{GAMI-Net local interpretation of a sample point in the bank marketing dataset.}\label{GAMINet:bank_local}	
\end{figure}
Due to the limit of page size, we only present the top 6 main effects and top 3 pairwise interaction, which is a subset of the selected 11 (out of 16) main effects and 8 (out of 20) pairwise interactions. The top three important variables are ``duration'' (last contact duration, in seconds), ``contact'' (contact communication type), and ``month'' (last contact month of year). The two most significant pairwise interactions are ``day vs. month'' and ``housing vs. month'', which are the statistics of the last contact. The fitted results of EBM in Figure~\ref{GAMINet:bank_ebm} is rather difficult to understand. For instance, there exist significant fluctuations as the variable ``age'' is greater than 80. It is hard to explain why the predicted outcomes of a client change greatly when his age increases from 80 to 90. In contrast, GAMI-Net is free from these problems as its fitted model is continuous and even smooth. Finally, the local interpretability of GAMI-Net is demonstrated in Figure~\ref{GAMINet:bank_local}, which shows the prediction diagnosis of one sample point.

\begin{table}[!t]
	\centering
	\renewcommand\tabcolsep{3pt}
	\renewcommand\arraystretch{1.2}
	\caption{Comparison results of the bank marketing dataset.} \label{GAMINet:Bank_Ablation}
	\begin{tabular}{c|c|cccc}
		\hline
		\multicolumn{2}{c|}{Model}                    &   Train AUC   &    Val AUC    &   Test AUC    &   Clarity Loss    \\ \hline
		\multicolumn{2}{c|}{XGBoost}  & 97.38$\pm$0.09 & 93.24$\pm$0.52 & 93.01$\pm$0.26  &         -    \\
		\multicolumn{2}{c|}{RF}                      & 93.59$\pm$0.10 & 92.01$\pm$0.62 & 91.98$\pm$0.35 &         -         \\
		\multicolumn{2}{c|}{MLP}                     & 93.15$\pm$0.65 & 93.12$\pm$0.90 & 92.29$\pm$0.41 &         -         \\	\multicolumn{2}{c|}{GLM}                     & 90.81$\pm$0.21 & 90.77$\pm$0.47 & 90.60$\pm$0.30 &         -         \\
		\multicolumn{2}{c|}{pyGAM}                    & 91.90$\pm$0.17 & 91.68$\pm$0.39 & 91.53$\pm$0.36 &         -         \\
		\multicolumn{2}{c|}{EBM}                     & 94.59$\pm$0.16 & 94.63$\pm$0.40 & 93.16$\pm$0.31 & 1.0228$\pm$0.0755 \\ \hline
		\multirow{10}{*}{\shortstack{G\\A\\M\\I\\N\\E\\T}} &   $\lambda=10^{0}$   & 92.15$\pm$0.42 & 92.08$\pm$0.46 & 91.85$\pm$0.42 & 0.0015$\pm$0.0016 \\	
		& $\lambda=8\times10^{-1}$  & 92.38$\pm$0.43 & 92.30$\pm$0.42 & 92.09$\pm$0.40 & 0.0018$\pm$0.0022 \\
		& $\lambda=6\times10^{-1}$ & 92.74$\pm$0.17 & 92.67$\pm$0.44 & 92.42$\pm$0.30 & 0.0038$\pm$0.0022 \\
		& $\lambda=4\times10^{-1}$ & 93.01$\pm$0.17 & 92.91$\pm$0.48 & 92.65$\pm$0.36 & 0.0052$\pm$0.0038 \\
		& $\lambda=2\times10^{-1}$ & 93.46$\pm$0.23 & 93.25$\pm$0.44 & 93.04$\pm$0.31 & 0.0112$\pm$0.0044 \\	
		&  $\lambda=10^{-1}$   & 93.94$\pm$0.23 & 93.52$\pm$0.41 & 93.33$\pm$0.39 & 0.0284$\pm$0.0138 \\
		&  $\lambda=10^{-2}$   & 94.22$\pm$0.20 & 93.53$\pm$0.38 & 93.38$\pm$0.33 & 0.0737$\pm$0.0255 \\
		&  $\lambda=10^{-3}$   & 94.25$\pm$0.15 & 93.54$\pm$0.40 & 93.36$\pm$0.32 & 0.3733$\pm$0.0785 \\
		&  $\lambda=10^{-4}$   & 94.19$\pm$0.26 & 93.52$\pm$0.42 & 93.34$\pm$0.30 & 1.5088$\pm$0.2752 \\
		\cline{2-6}
		&  \tabincell{c}{No Heredity\\ $\lambda=10^{-1}$} & 93.94$\pm$0.23  & 93.51$\pm$0.41 & 93.33$\pm$0.40 & 0.0305$\pm$0.0166 \\ \hline\hline
	\end{tabular}
\end{table}
The comparison results of different methods are shown in Table~\ref{GAMINet:Bank_Ablation}, together with the GAMI-Net results under different settings. From the results, we can obtain similar conclusions to that of the simulation study. With the increase of $\lambda$, the marginal clarity losses get decreased while the predictive performance does not change too much until $\lambda=10^{-1}$. Therefore, $\lambda$ is set to $10^{-1}$ for this dataset to pursue a good balance between model interpretability and predictive performance. In addition, it is observed that GAMI-Net with and without the heredity constraint provide the same results, which indicates that the model following heredity constraint is naturally favored regarding the predictive performance. And the use of heredity constraint can still help reduce the search space during model estimation.	

\subsection{Bike Sharing Hour Dataset}
The bike sharing hour dataset is a regression task with 17379 samples and 12 explanatory variables (\url{https://archive.ics.uci.edu/ml/datasets/bike+sharing+dataset}). Each sample records the basic environmental information including 8 categorical variables, e.g., the season and the weather situation; and 4 numerical variables, e.g., the temperature and wind speed. The target is to predict the hourly count of rental bikes in the capital bike share system between 2011 and 2012.

\begin{figure*}[!t]
	\centering
	\subfloat[GAMI-Net]{
		\label{GAMINet:bike_share_gaminet} 
		\includegraphics[width=0.475\textwidth]{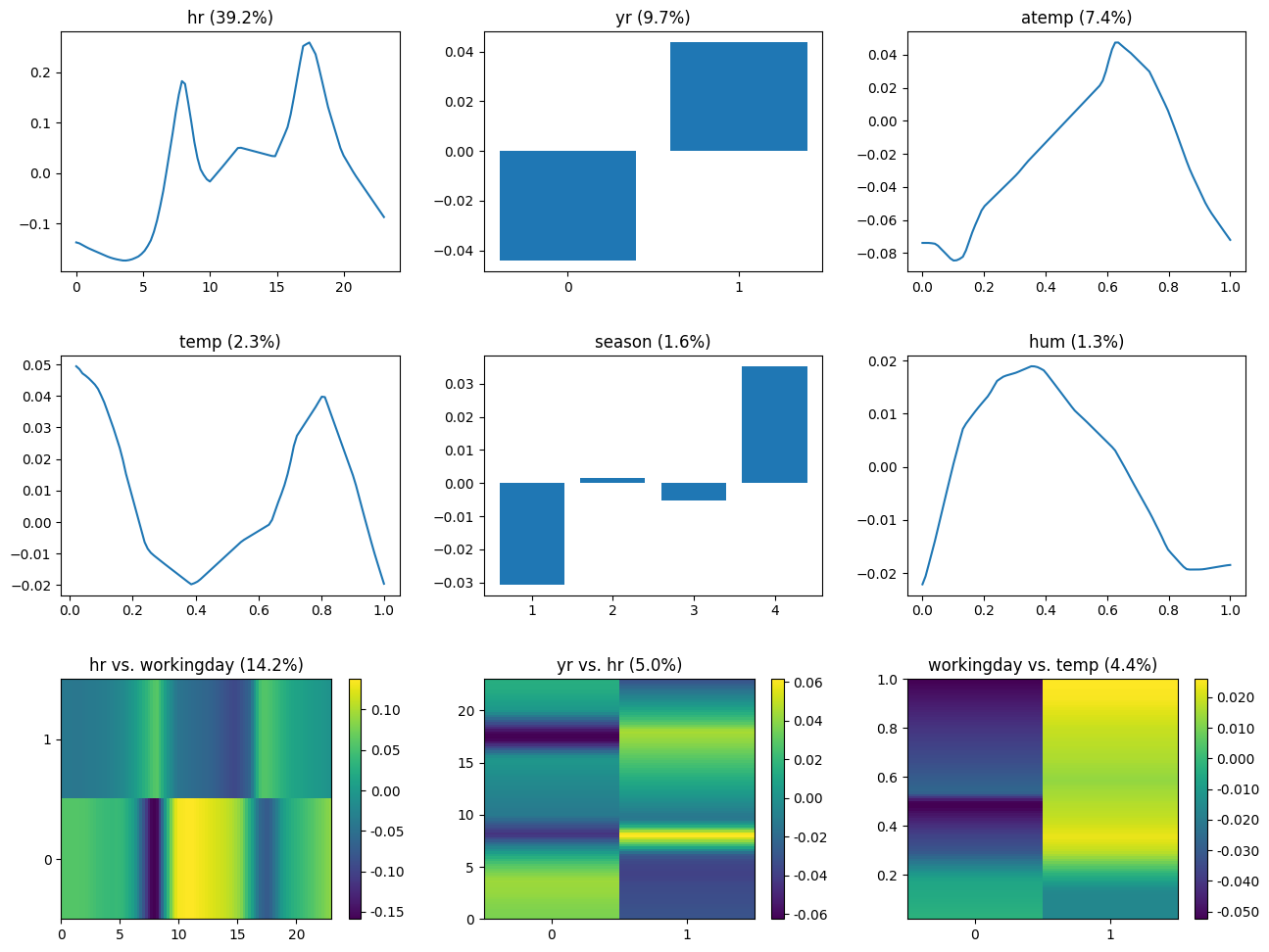}}
	\quad
	\subfloat[EBM]{
		\label{GAMINet:bike_share_ebm} 
		\includegraphics[width=0.475\textwidth]{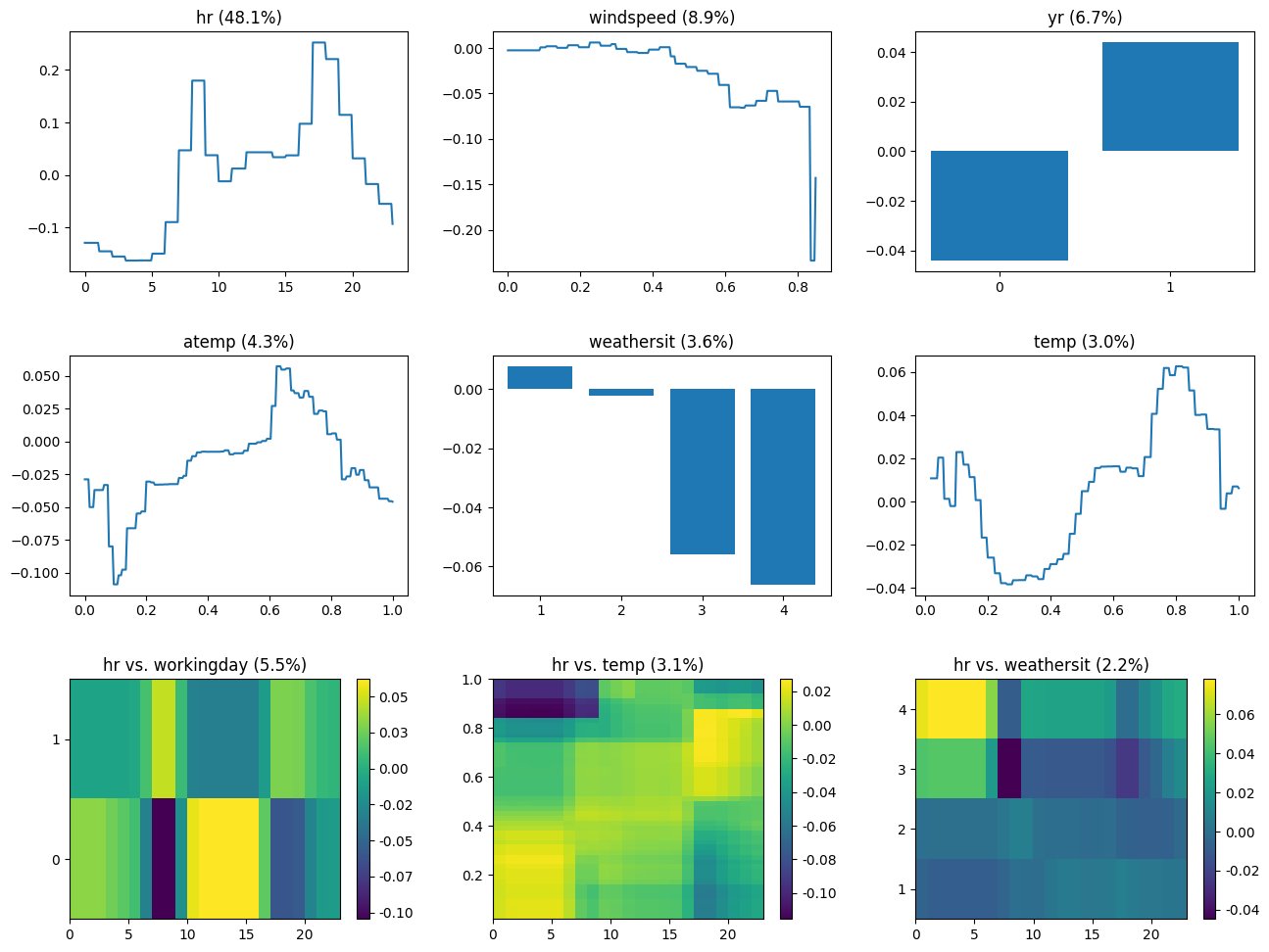}} 
	\caption{GAMI-Net vs. EBM global interpretation for the bike sharing hour dataset.}\label{GAMINet:bike_share_global}	
\end{figure*}

Similarly, the global interpretation of the bike sharing hour dataset is shown in Figure~\ref{GAMINet:bike_share_global}. In total, 9 (out of 12) main effects and all the 20 pairwise interactions are shown to be non-trivial. The most important variable is ``hr'' (hour, ranges from 0 to 23) with IR equals to 39.2\%. It can be observed that there exist two peaks of bike sharing around 8 AM and 5 PM, which correspond to the rush hour in a day. The categorical variable ``yr'' (year, 0 denotes 2011 and 1 means 2012) is the second important one, and the results show that there exists an increasing trend of bike sharing over time. The third important variable is ``atemp'' (feeling temperature in Celsius divided by 50), and low feeling temperature (below 20 Celsius) is a negative factor for bike sharing. Regarding pairwise interactions, the ``hr vs. workingday'' is shown to be relatively important.

\begin{table}
	\centering
	\renewcommand\tabcolsep{3pt}
	\renewcommand\arraystretch{1.2}
	\caption{Comparison results of the bike sharing hour dataset.} \label{GAMINet:Bike_Ablation}
	\begin{tabular}{c|c|cccc}
		\hline
		\multicolumn{2}{c|}{Model}                           &   Train RMSE    &    Val RMSE     &    Test RMSE    &   Clarity Loss    \\ \hline
		\multicolumn{2}{c|}{XGBoost}                          &  5.64$\pm$4.52  & 40.96$\pm$1.17  & 42.33$\pm$0.72  &         -         \\
		\multicolumn{2}{c|}{RF}                            & 62.00$\pm$0.72  & 65.12$\pm$1.47  & 65.69$\pm$2.10  &         -         \\
		\multicolumn{2}{c|}{MLP}                            & 35.54$\pm$1.31  & 37.09$\pm$1.52  & 43.35$\pm$1.32  &         -         \\
		\multicolumn{2}{c|}{GLM}                            & 158.98$\pm$0.52 & 158.54$\pm$2.49 & 159.09$\pm$1.57 &         -         \\
		\multicolumn{2}{c|}{pyGAM}                           & 99.52$\pm$0.60  & 100.16$\pm$1.61 & 100.22$\pm$1.12 &         -         \\
		\multicolumn{2}{c|}{EBM}                            & 54.32$\pm$0.55  & 54.27$\pm$0.99  & 57.48$\pm$1.20  & 0.0026$\pm$0.0005 \\ \hline
		\multirow{10}{*}{\shortstack{G\\A\\M\\I\\N\\E\\T}} &     $\lambda=10^{0}$     & 52.93$\pm$0.80  & 55.02$\pm$1.23  & 55.81$\pm$1.13  & 0.0005$\pm$0.0001 \\
		& $\lambda=8\times10^{-1}$ & 52.12$\pm$0.93  & 54.45$\pm$1.23  & 54.99$\pm$1.00  & 0.0005$\pm$0.0001 \\
		& $\lambda=6\times10^{-1}$ & 51.26$\pm$0.80  & 53.63$\pm$1.11  & 54.27$\pm$1.05  & 0.0006$\pm$0.0001 \\
		& $\lambda=4\times10^{-1}$ & 50.84$\pm$1.07  & 53.31$\pm$1.03  & 53.91$\pm$1.31  & 0.0007$\pm$0.0002 \\
		& $\lambda=2\times10^{-1}$ & 50.21$\pm$0.96  & 52.98$\pm$1.32  & 53.47$\pm$1.19  & 0.0007$\pm$0.0001 \\
		&    $\lambda=10^{-1}$     & 50.26$\pm$1.33  & 53.38$\pm$1.85  & 53.68$\pm$1.83  & 0.0007$\pm$0.0001 \\
		&    $\lambda=10^{-2}$     & 50.00$\pm$0.93  & 53.02$\pm$1.33  & 53.40$\pm$1.12  & 0.0014$\pm$0.0002 \\
		&    $\lambda=10^{-3}$     & 50.22$\pm$0.95  & 53.08$\pm$1.53  & 53.71$\pm$1.07  & 0.0050$\pm$0.0033 \\
		&    $\lambda=10^{-4}$     & 50.08$\pm$1.08  & 53.03$\pm$0.92  & 53.62$\pm$0.96  & 0.0137$\pm$0.0067 \\ \cline{2-6}
		&  \tabincell{c}{No Heredity\\ $\lambda=10^{-1}$}   & 50.26$\pm$1.33  & 53.38$\pm$1.85  & 53.68$\pm$1.83  & 0.0007$\pm$0.0001 \\ \hline\hline
	\end{tabular}
\end{table}

The comparison results of the bike sharing hour dataset are shown in Table~\ref{GAMINet:Bike_Ablation}, in which consistent conclusions can be derived. The default marginal clarity regularization strength is also set to $10^{-1}$ considering the balance between predictive performance and model interpretability. 

\subsection{More Real-world Datasets}
In addition to the simulation study and 2 real-world applications, we test the predictive performance of the proposed GAMI-Net on another 20 regression datasets, collected from different domains. These datasets are available in the UCI machine learning repository or the OpenML platform. The sample sizes range from 500 (no2) to 20640 (california housing) and the number of variables also varies from 3 (disclosure z) to 266 (topo 2 1). The detailed information of each dataset is presented in Table~\ref{GAMINet:real_reg_rmse}, where the test set performance of each compared method is reported. Note all the listed results should be multiplied by the corresponding scaling factors in the last column.

\begin{sidewaystable}[!t]
	\centering
	\small
	\renewcommand\tabcolsep{1.5pt}
	\renewcommand\arraystretch{1.1}
	\caption{Test set RMSE on 20 real-world regression datasets.} \label{GAMINet:real_reg_rmse}
	\begin{tabular}{c|cc|cccc|ccc|c}
		\hline
		Dataset       &  $n$  & $p$ &          GAMI-Net          &            EBM             &           pyGAM            &       GLM       &            MLP             &             RF             &          XGBoost           &       Scale        \\ \hline
		no2         &  500  &  7  &      4.954$\pm$0.444       & $\mathbf{4.681}$$\pm$0.396 &      4.971$\pm$0.514       & 6.508$\pm$0.626 &      5.201$\pm$0.466       & $\mathbf{4.688}$$\pm$0.422 &      4.911$\pm$0.342       &  $\times  0.  1$   \\
		sensory       &  576  & 11  &      7.335$\pm$0.541       & $\mathbf{7.054}$$\pm$0.475 &      7.923$\pm$0.279       & 8.066$\pm$0.226 &      7.455$\pm$0.339       & $\mathbf{7.318}$$\pm$0.497 &      8.205$\pm$0.549       &  $\times  0.  1$   \\
		disclosure z    &  662  &  3  & $\mathbf{2.420}$$\pm$0.275 &      2.438$\pm$0.275       &      2.429$\pm$0.277       & 2.438$\pm$0.267 & $\mathbf{2.442}$$\pm$0.270 &      2.445$\pm$0.260       &      2.856$\pm$0.184       &  $\times  10000$   \\
		bike sharing day   &  731  & 11  &      0.691$\pm$0.029       & $\mathbf{0.663}$$\pm$0.043 &      0.710$\pm$0.036       & 1.144$\pm$0.045 &      0.827$\pm$0.042       &      0.727$\pm$0.051       & $\mathbf{0.711}$$\pm$0.067 &   $\times  1000$   \\
		era         & 1000  &  4  & $\mathbf{1.566}$$\pm$0.037 &      1.566$\pm$0.038       &      1.568$\pm$0.039       & 1.684$\pm$0.041 &      1.583$\pm$0.041       & $\mathbf{1.573}$$\pm$0.041 &      1.596$\pm$0.045       &    $\times  1$     \\
		treasury      & 1049  & 15  &      2.187$\pm$0.268       &      2.513$\pm$0.400       & $\mathbf{2.114}$$\pm$0.260 & 8.971$\pm$0.739 & $\mathbf{2.367}$$\pm$0.282 &      2.416$\pm$0.324       &      2.489$\pm$0.247       &  $\times  0.  1$   \\
		weather izmir    & 1461  &  9  & $\mathbf{1.105}$$\pm$0.128 &      1.322$\pm$0.073       &      1.150$\pm$0.131       & 3.231$\pm$0.133 & $\mathbf{1.289}$$\pm$0.129 &      1.303$\pm$0.103       &      1.337$\pm$0.095       &    $\times  1$     \\
		airfoil       & 1503  &  5  & $\mathbf{2.080}$$\pm$0.141 &      2.169$\pm$0.100       &      4.563$\pm$0.194       & 6.357$\pm$0.205 &      2.607$\pm$0.269       &      2.440$\pm$0.126       & $\mathbf{1.742}$$\pm$0.161 &    $\times  1$     \\
		wine red      & 1599  & 11  &      6.219$\pm$0.147       & $\mathbf{5.991}$$\pm$0.231 &      6.252$\pm$0.129       & 7.473$\pm$0.180 &      6.201$\pm$0.165       & $\mathbf{5.918}$$\pm$0.210 &      6.135$\pm$0.251       &  $\times  0.  1$   \\
		skill craft     & 3395  & 18  &      0.970$\pm$0.025       & $\mathbf{0.920}$$\pm$0.026 &      1.154$\pm$0.531       & 1.200$\pm$0.022 &      1.019$\pm$0.075       & $\mathbf{0.927}$$\pm$0.025 &      0.997$\pm$0.021       &    $\times  1$     \\
		abalone       & 4177  &  8  & $\mathbf{2.132}$$\pm$0.051 &      2.240$\pm$0.052       &      2.168$\pm$0.091       & 2.975$\pm$0.126 & $\mathbf{2.142}$$\pm$0.078 &      2.186$\pm$0.074       &      2.372$\pm$0.066       &    $\times  1$     \\
		parkinsons tele   & 5875  & 19  & $\mathbf{0.364}$$\pm$0.028 &      0.412$\pm$0.008       &      0.771$\pm$0.035       & 1.061$\pm$0.018 &      0.580$\pm$0.026       &      0.321$\pm$0.014       & $\mathbf{0.198}$$\pm$0.009 &    $\times  10$    \\
		wind        & 6574  & 14  & $\mathbf{3.050}$$\pm$0.064 &      3.085$\pm$0.064       &      3.071$\pm$0.062       & 4.590$\pm$0.064 & $\mathbf{3.046}$$\pm$0.071 &      3.258$\pm$0.069       &      3.205$\pm$0.095       &    $\times  1$     \\
		cpu small      & 8192  & 12  &      0.292$\pm$0.008       & $\mathbf{0.286}$$\pm$0.010 &      0.327$\pm$0.011       & 1.464$\pm$0.049 &      0.310$\pm$0.006       &      0.314$\pm$0.011       & $\mathbf{0.294}$$\pm$0.023 &    $\times  10$    \\
		topo 2 1      & 8885  & 266 &      2.897$\pm$0.306       & $\mathbf{2.873}$$\pm$0.312 &      3.054$\pm$0.337       & 2.940$\pm$0.318 &      2.892$\pm$0.314       & $\mathbf{2.864}$$\pm$0.318 &      3.067$\pm$0.301       &  $\times  0.0  1$  \\
		ccpp        & 9568  &  4  &      3.873$\pm$0.063       & $\mathbf{3.673}$$\pm$0.069 &      4.087$\pm$0.084       & 6.143$\pm$0.053 &      4.157$\pm$0.084       &      3.705$\pm$0.073       & $\mathbf{3.121}$$\pm$0.090 &    $\times  1$     \\
		electrical grid   & 10000 & 11  & $\mathbf{0.927}$$\pm$0.021 &      0.951$\pm$0.016       &      1.718$\pm$0.020       & 2.885$\pm$0.049 & $\mathbf{0.658}$$\pm$0.019 &      1.448$\pm$0.028       &      0.993$\pm$0.016       &  $\times  0.0  1$  \\
		ailerons      & 13750 & 40  & $\mathbf{1.640}$$\pm$0.049 &      1.660$\pm$0.049       &      1.690$\pm$0.054       & 3.380$\pm$0.087 & $\mathbf{1.590}$$\pm$0.030 &      1.680$\pm$0.040       &      1.650$\pm$0.050       & $\times  0.000  1$ \\
		elevators      & 16599 & 18  & $\mathbf{2.253}$$\pm$0.079 &      2.284$\pm$0.087       &      2.389$\pm$0.053       & 6.710$\pm$0.167 & $\mathbf{2.106}$$\pm$0.070 &      3.147$\pm$0.064       &      2.167$\pm$0.041       & $\times  0.00  1$  \\
		california housing & 20640 &  8  & $\mathbf{5.181}$$\pm$0.072 &      5.235$\pm$0.084       &      6.504$\pm$0.327       & 9.160$\pm$0.091 &      5.777$\pm$0.169       &      5.796$\pm$0.095       & $\mathbf{4.708}$$\pm$0.087 &  $\times  0.  1$   \\ \hline\hline
	\end{tabular}
\end{sidewaystable}

Generally speaking, GAMI-Net shows comparative predictive performance to that of EBM and other benchmark models. In practice, it is hard to say which one will perform better. GAMI-Net is more likely to have better predictive performance when the actual shape functions are continuous and smooth. In contrast, EBM could perform better as the shape functions are piece-wise constant. Both of them are competitive regarding predictive power, while the proposed GAMI-Net is designed with more interpretability considerations. Therefore, GAMI-Net is a promising tool in the area of interpretable machine learning.

\section{Conclusion} \label{Conclusion}
In this paper, an intrinsically explainable neural network called GAMI-Net is proposed. It approximates the complex functional relationship using subnetwork-represented main effects and pairwise interactions, which can be easily interpreted using 1D line plots / bar charts and 2D heatmaps. Several statistically meaningful constraints are considered to enhance the model interpretability, including the heredity constraint for enforcing structural pairwise interactions, the sparsity constraint for promoting model parsimony, and the marginal clarity constraint for avoiding effects mixing problem. The experimental results show that the proposed model has competitive predictive performance to black-box machine learning models. Meanwhile, the model estimated by GAMI-Net is highly interpretable and easily visualized.

Some future works can be done for extending GAMI-Net. One direction is to consider additional shape constraints for each component function, e.g., monotonic increasing/decreasing, convex or concave, according to prior experience or domain knowledge. Another direction is to consider higher-order interactions for more sophisticated developments. 

\bibliographystyle{plainnat}
\bibliography{mybibfile}

\end{document}